\documentclass[runningheads]{llncs}
\usepackage{graphicx}
\usepackage{amsmath,amssymb,amsfonts}
\usepackage{float}
\usepackage{subcaption}
\usepackage{algorithm}
\usepackage{algpseudocode}
\algrenewcommand\algorithmicrequire{\textbf{Input:}}
\algrenewcommand\algorithmicensure{\textbf{Output:}}

\DeclareMathOperator{\xx}{\mathbf{x}}
\DeclareMathOperator{\zz}{\mathbf{z}}
\DeclareMathOperator{\cc}{\mathbf{c}}

\usepackage{multirow}
\usepackage{color,soul}

\usepackage[misc]{ifsym}

\begin{document}
\title{Generative Adversarial Networks for\\Failure Prediction}

\titlerunning{GANs for Failure Prediction}

\author{Shuai Zheng\textsuperscript{(\Letter)} \and Ahmed Farahat \and Chetan Gupta}

\authorrunning{S. Zheng et al.}
\institute{Industrial AI Lab, Hitachi America Ltd\\Santa Clara, CA, USA\\
\email{\{Shuai.Zheng,Ahmed.Farahat,Chetan.Gupta\}@hal.hitachi.com}}

\tocauthor{Shuai Zheng, Ahmed Farahat, Chetan Gupta}
\toctitle{Generative Adversarial Networks for Failure Prediction}

%Princeton University, Princeton NJ 08544, USA \and
%Springer Heidelberg, Tiergartenstr. 17, 69121 Heidelberg, Germany
\maketitle              % typeset the header of the contribution
\begin{abstract}
Prognostics and Health Management (PHM) is an emerging engineering discipline which is concerned with the analysis and prediction of equipment health and performance. One of the key challenges in PHM is to accurately predict impending failures in the equipment. In recent years, solutions for failure prediction have evolved from building complex physical models to the use of machine learning algorithms that leverage the data generated by the equipment. However, failure prediction problems pose a set of unique challenges that make direct application of traditional classification and prediction algorithms impractical. These challenges include the highly imbalanced training data, the extremely high cost of collecting more failure samples, and the complexity of the failure patterns. Traditional oversampling techniques will not be able to capture such complexity and accordingly result in overfitting the training data. This paper addresses these challenges by proposing a novel algorithm for failure prediction using Generative Adversarial Networks (GAN-FP). GAN-FP first utilizes two GAN networks to simultaneously generate training samples and build an inference network that can be used to predict failures for new samples. GAN-FP first adopts an infoGAN to generate realistic failure and non-failure samples, and initialize the weights of the first few layers of the inference network. The inference network is then tuned by optimizing a weighted loss objective using only real failure and non-failure samples. The inference network is further tuned using a second GAN whose purpose is to guarantee the consistency between the generated samples and corresponding labels. GAN-FP can be used for other imbalanced classification problems as well. Empirical evaluation on several benchmark datasets demonstrates that GAN-FP significantly outperforms existing approaches, including under-sampling, SMOTE, ADASYN, weighted loss, and infoGAN augmented training.    
\end{abstract}
\section{Introduction}

Reliability of industrial systems, products and equipment is critical not only to manufacturers, operating companies, but also to the entire society. For example, in 2017, due to electrical fault in a refrigerator, Grenfell Tower fire in London killed 72 people, hospitalized 74 and caused 200 million to 1 billion GBP property damage \cite{fire}. 
Because of the profound impact and extreme costs associated with system failures, methods that can predict and prevent such catastrophes have long been investigated. These methodologies can be grouped under the framework of Prognostics and Health Management (PHM), where prognostics is the process of predicting the future reliability of a product by assessing the extent of deviation or degradation of the product while the product is still working properly; health management is the process of real time measuring and monitoring. The benefits of accurate PHM approaches include: 1) providing advance warning of failures; 2) minimizing unnecessary maintenance, extending maintenance cycles, and maintaining effectiveness through timely repair actions; 3) reducing cost related to inspection and maintenance, reducing cost related to system downtime and inventory by scheduling replacement parts in the right time; and 4) improving the design of future systems \cite{vichare2006prognostics,mosallam2016data}. Failure prediction is one of the main tasks in PHM.

Failure prediction approaches can be categorized into model-based approaches and data-driven approaches \cite{pecht2010prognostics}. Model-based approaches use mathematical equations to incorporate a physical understanding of the system, and include both system modeling and physics-of-failure (PoF) modeling. The limitation is that the development of these models requires detailed knowledge of the underlying physical processes that lead to system failure. Furthermore, the physical models are often unable to model environmental interactions. Alternatively, data-driven techniques are gaining popularity. There are several reasons: 1) data-driven methods learn the behavior of the system based on monitored data and can work with incomplete domain knowledge; 2) data-driven methods can learn correlations between parameters and work well in complex systems, such as aircraft engines, HPC systems \cite{zheng2011analysis}, large manufacturing systems; 3) with the development of IoT systems, large amount of data is being collected in real time, which makes real time monitoring and alerts for PHM possible. 

However, data-driven techniques for failure prediction have a set of unique challenges. Firstly, for many systems and components, there is not enough failure examples in the training data. Physical equipment and systems are engineered not to fail and as a result failure data is rare and difficult to collect. Secondly, complex physical systems have multiple failure and degradation modes, often depending upon varying operating conditions. One way to overcome these challenges is to artificially generate failure data such that different failure modes and operating conditions are adequately covered and machine learning models can be learned over this augmented data. Traditionally, oversampling has been used to generate more training samples. However, oversampling cannot capture the complexity of the failure patterns and can easily introduce undesirable noise with overfitting risks due to the limitation of oversampling models. With the successful application of Generative Adversarial Networks (GANs) \cite{goodfellow2014generative} in other domains, GANs provide a natural way to generate additional data. For example, in computer vision, GANs are used to generate realistic images to improve performance in applications, such as, biomedical imaging \cite{calimeri2017biomedical}, person re-identification \cite{zhong2018camera} and image enhancement \cite{yun2018predicting}. In addition, GANs have been used to augment classification problems by using semi-supervised learning \cite{dai2017good,tran2017bayesian} or domain adaptation \cite{bousmalis2017unsupervised}. However, GAN methods cannot guarantee the consistency of the generated samples and their corresponding labels. For example, infoGAN is claimed to have $5\%$ error rate in generating MNIST digits \cite{chen2016infogan}.

In this work, we propose a novel algorithm that utilizes GANs for Failure Prediction (GAN-FP). Compared to existing work, our contributions are:
\begin{enumerate}
  \item We propose GAN-FP in which three different modules work collaboratively to train an inference network: (1) In one module, realistic failure and non-failure samples are generated using infoGAN. (2) In another module, the weighted loss objective is adopted to train inference network using real failure and non-failure samples. In our design, this inference network shares the weights of the first few layers with the discriminator network of the infoGAN. (3) In the third module, the inference network is further tuned using a second GAN by enforcing consistency between the output of the first GAN and label generated by the inference network. 
  \item We design a collaborative mini-batch training scheme for GANs and the inference network;
  \item We conduct several experiments that show significant improvement over existing approaches according to different evaluation criteria. Through visualization, we verify that GAN-FP generates realistic sensor data, and captures discriminative features of failure and non-failure samples. 
\item Failure prediction is the motivation and typical use case of our design. For broader applications, GAN-FP can be applied to other general imbalanced classification problems as well.
\end{enumerate}

\section{Background}

\subsection{Imbalanced classification for failure prediction}
Existing approaches to handle imbalanced data can be categorized into two groups: re-sampling (oversampling/undersampling) and cost-sensitive learning. \textbf{Re-sampling} method aims to balance the class priors by undersampling the majority non-failure class or oversampling the minority failure class (or both) \cite{nejatian2018using}. Chawla \textit{et al.} \cite{chawla2002smote} proposed SMOTE oversampling, which generates new synthetic examples from the minority class between the closest neighbors from this class. He \textit{et al.} \cite{he2008adasyn} proposed ADASYN oversampling, which uses a weighted distribution for different minority class examples according to their level of difficulty in learning. Inspired by the success of boosting algorithms and ensemble learning, re-sampling techniques have been integrated into ensemble learning \cite{nejatian2018using}. \textbf{Cost-sensitive learning} assigns higher misclassification costs to the failure class than to the non-failure class \cite{shen2015deepcontour}. Zhang \textit{et al.} \cite{zhang2018cost} proposed an evolutionary cost-sensitive deep belief network for imbalanced classification, which uses adaptive differential evolution to optimize the misclassification costs. Using weighted softmax loss function, Jia \textit{et al.} \cite{jia2018deep} proposed a framework called Deep Normalized CNN for imbalanced fault classification of machinery to overcome data imbalanced distribution. Many hybrid methods combine both re-sampling and cost-sensitive learning \cite{tang2009svms}. However, limitations exist in both categories. For instance, oversampling can easily introduce undesirable noise with overfitting risks, and undersampling removes valuable information due to data loss. Cost-sensitive learning requires a good insight into the modified learning algorithms and a precise identification of reasons for failure in mining skewed distributions. Data with highly skewed classes also pose a challenge to traditional discriminant algorithms, such as subspace and feature representation learning \cite{zheng2014kernel,zheng2015closed,zheng2016harmonic,zheng2017machine,zheng2018harmonic,zheng2019sparse}. 
This makes it further difficult to achieve ideal classification accuracy in failure prediction tasks. 

\subsection{GAN}
Recently, generative models such as Generative Adversarial Networks (GANs) have attracted a lot of interest from researchers and industrial practitioners. Goodfellow \textit{et al.} formulated GAN into a minimax two-player game, where they simultaneously train two models: a generator network $G$ that captures the data distribution, and a discriminator network $D$ that estimates the probability that a sample comes from the true data rather than the generator network $G$. The goal is to learn the generator distribution $p(\xx')$ of $G$ over data $\xx'$ that matches the real data distribution $p(\xx)$. The generator network $G$ generates samples by transforming a noise vector $\zz \sim p(\zz)$ into a generated sample $G(\zz)$. This generator is trained by playing against the discriminator network $D$ that aims to distinguish between samples from true data distribution $p(\xx)$ and the generator distribution $p(\xx')$. Formally, the minimax game is given by the following expression:
\begin{align}
\min_G \max_D V(D,G) = \mathbb{E}_{\xx \sim p(\xx)}[\log D(\xx)] + \mathbb{E}_{\zz \sim p(\zz)} [\log (1-D(  G(\zz)  )  )].
\label{eq:gan}
\end{align}

\begin{figure}[t]
\centering
\includegraphics[width=0.6\textwidth]{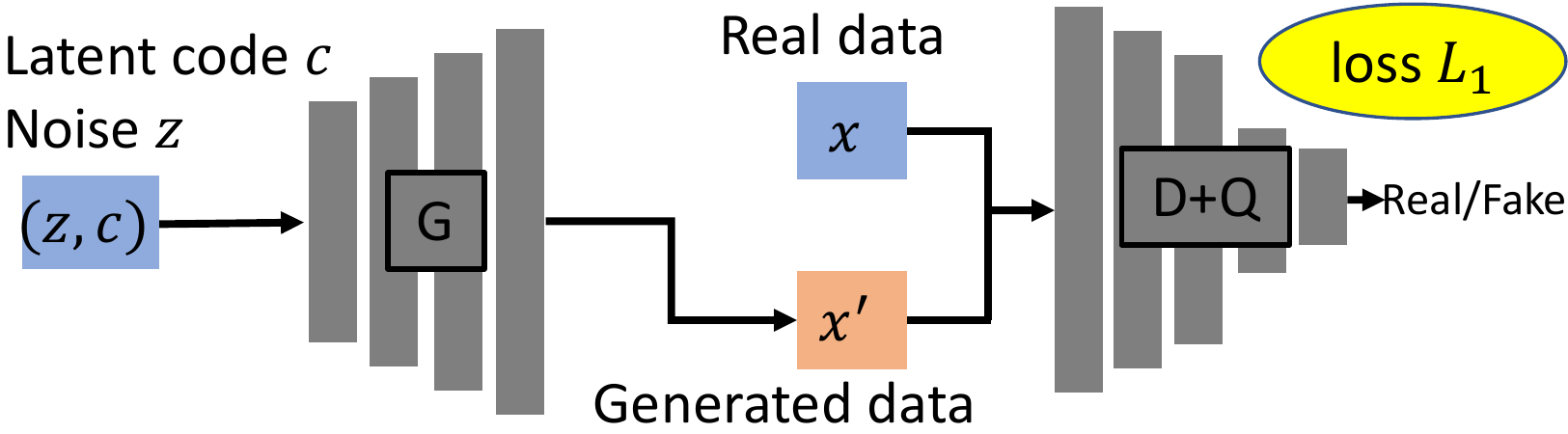}
\caption{InfoGAN architecture: latent code $\cc$ and noise vector $\zz$ are combined as input for generator $G$, $\xx' = G(\zz, \cc)$ is generated data, $\xx$ is real data, discriminator $D$ tries to distinguish generated data from real data, network $Q$ is used to maximize $L_{mutual}$ (Eq.(\ref{eq:mutual})), loss $L_1$ is given in Eq.(\ref{eq:infogan}).}
\label{fig:infogan}
\end{figure}

InfoGAN \cite{chen2016infogan} is an information-theoretic extension to GAN. InfoGAN decomposes the input noise vector into two parts: incompressible noise vector $\zz$ and latent code vector $\cc$. The latent code vector $\cc$ targets the salient structured semantic features of the data distribution and can be further divided into categorical and continuous latent code, where the categorical code controls sample labels and continuous code controls variations. Thus, in infoGAN, the generated sample becomes $G(\zz, \cc)$. InfoGAN introduces a distribution $Q(\cc|\xx)$ to approximate $p(\cc|\xx)$ and maximizes the variational lower bound, $L_{mutual}(G,Q)$, of the mutual information, $I(\cc;G(\zz, \cc))$:
\begin{align}
L_{mutual}(G,Q) = \mathbb{E}_{\cc \sim p (\cc), \xx \sim G(\zz, \cc)}[\log Q(\cc | \xx)] + H(c),
\label{eq:mutual}
\end{align}
where $L_{mutual}(G,Q)$ is easy to approximate with Monte Carlo simulation. In practice, $L_{mutual}$ can be maximized with respect to $Q$ directly and with respect to $G$ via the reparametrization trick. InfoGAN is defined as the following minimax game with a variational regularization of mutual information and a hyperparameter $\lambda_Q$:
\begin{align}
\min_{G,Q} \max_D L_1(D, G, Q) = V(D,G) - \lambda_Q L_{mutual}(G,Q).
\label{eq:infogan}
\end{align}
Figure \ref{fig:infogan} shows the structure of infoGAN. 
% In most experiments, $Q$ and $D$ share the first several convolutional layers. 

\begin{figure}[t]
\centering
\includegraphics[width=0.6\textwidth]{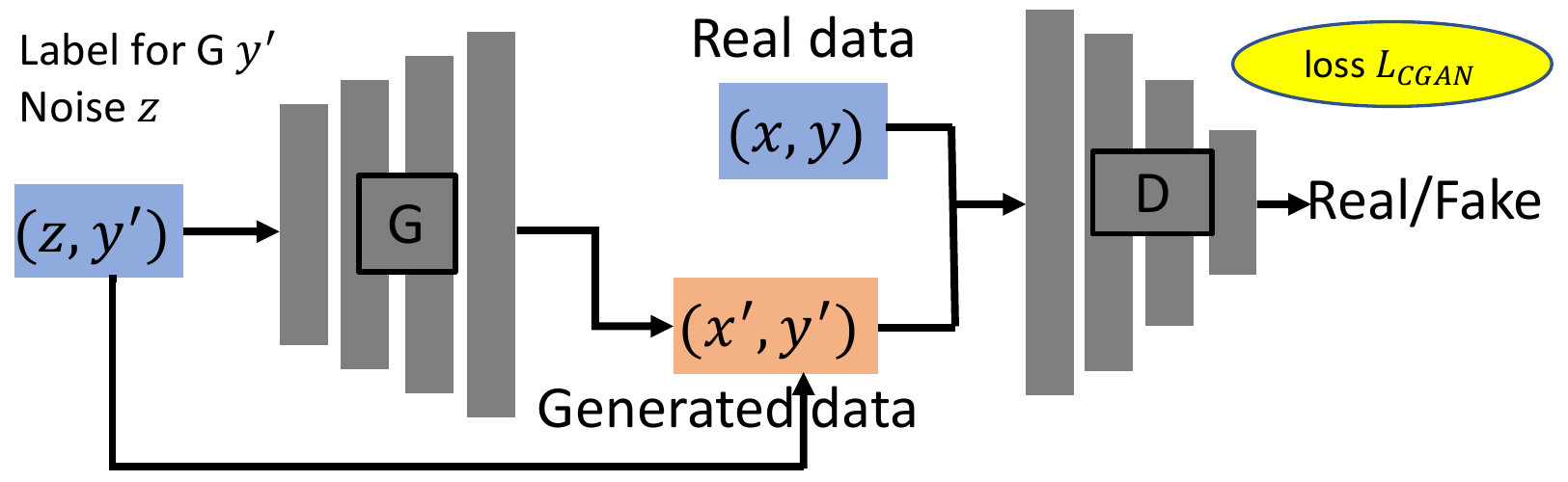}
\caption{Conditional GAN (CGAN) architecture: $y'$ and noise vector $\zz$ are combined as input for generator $G$, $\xx' = G(\zz, y')$ is generated data, $(\xx,y)$ is real data-label pair, $(  \xx', y'  )$ is generated data-label pair, discriminator $D$ tries to distinguish generated data-label pair from real data-label pair, loss $L_{CGAN}$ is given in Eq.(\ref{eq:cgan}).}
\label{fig:cgan}
\end{figure}

Conditional GAN (CGAN) \cite{mirza2014conditional} adds extra label information $y'$ to generator $G$ for conditional generation. In discriminator $D$, both $\xx$ and $y$ are presented as inputs and $D$ tries to distinguish if data-label pair is from generated or real data. 
Figure \ref{fig:cgan} shows the architecture of CGAN. The objective of CGAN is given as the following minimax game:
\begin{align}
\min_{G} \max_D L_{CGAN}(D, G) = &\mathbb{E}_{(\xx,y) \sim p(\xx,y)}[\log D(\xx, y)] + \nonumber \\
&
\mathbb{E}_{\zz \sim p(\zz)} [\log (1-D(  G(\zz, y'), y'  )  )].
\label{eq:cgan}
\end{align}

\section{GAN-FP: GAN for Failure Prediction}

\subsection{Motivation}
In failure prediction problems, we collect a lot of training data $\xx$ and the corresponding labels $y$. Training data $\xx$ usually is sensor data coming from equipment, but can be image, acoustics data as well. Label $y$ contains a lot of non-failure label $0$\textit{s} and very few failure label $1$\textit{s}.

Given a failure prediction problem, one choice is to construct a deep inference neural network and adopt weighted loss objective. As there are not enough real failure samples, test samples with failure labels are often misclassified to the prevalent non-failure class. As mentioned earlier, in this work, we propose the use of GANs to generate realistic failure samples.

To control the class labels of generated samples, we can choose Conditional GAN (CGAN) or infoGAN. CGAN was shown to mainly capture class-level features \cite{chen2016infogan,lee2017controllable}. In addition to capturing class-level features, infoGAN captures fine variations of features that are continuous in nature using continuous latent code.  As mentioned earlier, PHM data has multiple failure modes, and is continuous in nature, hence we use infoGAN as a basic building block in our design.

One problem with simply using infoGAN, is that it cannot guarantee that the generated sample is from a desired class. This means that some generated samples might end up having the wrong label. For example, infoGAN is claimed to have $5\%$ error rate in generating MNIST digits \cite{chen2016infogan}. When we have a 2-class highly imbalanced classification problem like failure prediction, this can have significant negative impact on the usefulness of this approach. In order to alleviate this problem, we propose the use of a second GAN to enforce the consistency of data-label pairs. In the second GAN, we use the inference network $P$ as a label generator. 

Once we have generated data, a traditional approach is to use both the generated and real samples to train a classifier. However, since we are sharing layers between the inference network and the discriminator network in the first GAN, and training all three modules simultaneously, we can directly use this inference network to achieve higher inference accuracy. 
 
During building the model, we alternate between the following steps:
\begin{enumerate}
\item Update the infoGAN to generate realistic samples for failure and non-failure labels.
\item Update the inference network $P$ using real data. We bootstrap $P$ using the weights of the first few layers of the discriminator of the infoGAN. This is a common approach to save training time and utilize the ability of GAN to extract features. 
\item Update inference network $P$ along with the discriminator in the second GAN to make sure that the generated samples and corresponding labels are consistent. This will increase the discriminative power of inference network $P$.
\end{enumerate}

\subsection{GAN-FP model}

\begin{figure}[t]
\centering
\includegraphics[width=1\textwidth]{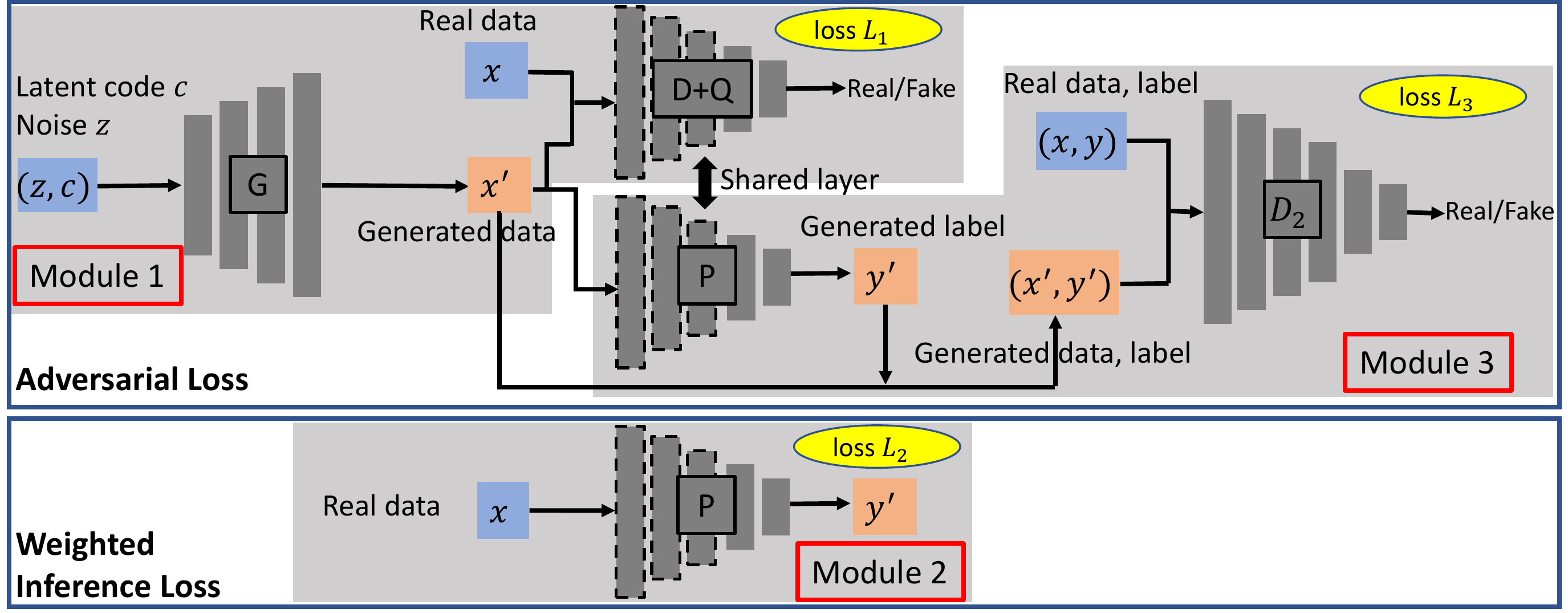}
\caption{GAN-FP architecture: there are 3 modules. Module 1 (network $G$, $D$ and $Q$) is used to generate failure and non-failure samples using adversarial loss $L_1$ (Eq.(\ref{eq:infogan})). Module 2 (network $P$) is an inference module with weighted loss $L_2$ (Eq.(\ref{eq:md2})), which trains a deep neural network using real data and label. Module 3 (network $P$ and $D_2$) is a modified CGAN module with adversarial loss $L_3$ (Eq.(\ref{eq:md3})), where network $D_2$ takes data-label pair as input and tries to distinguish whether the pair comes from real data label $(\xx,y)$ or from generated data label $(\xx',y')$. 
}
\label{fig:fpgan}
\end{figure}

Figure \ref{fig:fpgan} shows the design of GAN-FP. 

\textbf{Module 1} adopts an infoGAN to generate class-balanced samples. For the input categorical latent code $\cc$, we randomly generate labels $0$s (non-failure) and $1$s (failure) with equal probability. The continuous latent code $\cc$ and noise vector $\zz$ is generated using uniform random process within range $[0,1]$. 
Generator network $G$ is a deep neural network with input $(\zz,\cc)$, and outputs generated sample $\xx'$, where $\xx'$ has the same size as real data $\xx$. Discriminator network $D$ aims to distinguish generated sample $\xx'$ from real sample $\xx$. Network $Q$ aims to maximize the mutual information between latent code $\cc$ and generated sample $\xx'$. By jointly training network $G$, $D$ and $Q$, module 1 solves the minimax problem denoted in Eq.(\ref{eq:infogan}). The first few layers of the discriminative layer $D+Q$ will capture a lot of implicit features about the data. In order to reduce the overall training time, we are going to reuse these weights while training the inference network in the Module 2.

\textbf{Module 2} consists of a deep neural network $P$ and solves a binary classification problem with weighted loss based on real data and real label. Network $P$ shares the first several layers with $D$ and takes as input real data $\xx$ and outputs a probability (denoted as $P(\xx)$) within range $[0,1]$ indicating the chance that $\xx$ is a failure sample. The real label is denoted as $y$ ($0$ or $1$). In our design, the loss function $L_2$ for module 2 is cross entropy:
\begin{align}
\min_P L_2(P) = \mathbb{E}_{(\xx,y) \sim p(\xx, y)} [- w y \log (P(\xx)) - (1-y) \log (1-P(\xx))],
\label{eq:md2}
\end{align}
where weight $w = \frac{\text{number of non-failure samples}}{\text{number of failure samples}}>1$. Note at this step, the input for network $P$ is class-imbalanced real data and labels. Loss $L_2$ is a weighted version which emphasizes more on failure sample prediction. In the training of Module 3, the weights of inference network $P$ will be further tuned using generated data and labels.

\textbf{Module 3} consists of network $P$ and $D_2$ and enforces generated data-label pair $(\xx', y')$ to look like real data-label pair $(\xx, y)$. $P$ serves as the generator network. Given $\xx'$, the generated label $y'=P(\xx')$ needs to be as correct as possible. $D_2$ tries to distinguish the generated data-label pair from real pair. The minimax objective for module 3 is given as:
\begin{align}
\min_{P} \max_{D_2} L_3(P, D_2) =& \mathbb{E}_{(\xx,y) \sim p(\xx,y)}[\log D_2(\xx, y)] \nonumber \\
&
+ \mathbb{E}_{\xx' \sim p(\xx')} [\log (1-D_2(  [\xx', P(\xx')] )  )].
\label{eq:md3}
\end{align}
While training this module, the weights of the inference network $P$ will be further tuned to increase the discrimination between failure and non-failure labels. The effectiveness of Module 3 to improve inference network $P$ will be validated by comparing the performance of GAN-FP with infoGAN augmented training (denoted as InfoGAN AUG in experiments), where we train the inference network $P$ with generated data without using Module 3.

\begin{algorithm}[t]
\footnotesize
\caption{Mini-batch SGD solving GAN-FP.}
\label{alg:alg1}
\begin{algorithmic}[1]
\Require Real data and label pairs $\{ \xx_i, y_i\}$, where $i=1,2,...,n$, hyperparameter $\lambda_G$, $\lambda_D$, $\lambda_P$, $\lambda_{D_2}$, $\lambda_{L_2}$, batch size $b$.
\Ensure Network parameters $\theta_G$, $\theta_D$, $\theta_Q$, $\theta_P$, $\theta_{D_2}$ for networks $G$, $D$, $Q$, $P$, $D_2$ respectively.
\State Initialize $\theta_G$, $\theta_D$, $\theta_Q$, $\theta_P$, $\theta_{D_2}$.
\Repeat 
\State Randomly choose $b$ data and label pairs from $\{ \xx_i, y_i\}$.
\State Randomly generate $b$ latent code $\cc$ and noise $\zz$, where $\cc$ is class-balanced, noise $\zz$ is uniform random variables. 
\State Update Module 1 discriminator network $\theta_D$ by ascending along its stochastic gradient w.r.t. $\max_D \lambda_D L_1(D)$ and share the weights of the first few layers with $P$.
\State Update Module 1 generator and $Q$-network $\theta_G$, $\theta_Q$ by descending along its stochastic gradient w.r.t. $\min_{G,Q} \lambda_G L_1(G, Q)$.
\State Update inference network $\theta_P$ by descending along its stochastic gradient w.r.t. $\min_P \lambda_{L_2} L_2(P)$ and use $P$ as the generator of Module 3.
\State Update Module 3 discriminator network $\theta_{D_2}$ by ascending along its stochastic gradient w.r.t. $\max_D \lambda_{D_2} L_3(D_2)$.
\State Update Module 3 generator network $\theta_P$ by descending along its stochastic gradient w.r.t. $\min_{P} \lambda_P L_3(P)$.
\Until{Convergence}
\end{algorithmic}
\end{algorithm}

\subsection{Algorithm}
Algorithm \ref{alg:alg1} summarizes the procedure for training GAN-FP. The input data includes real data-label pairs $(\xx_i, y_i)$, where $i=1,2,...,n$, and hyperparameter $\lambda_G$, $\lambda_D$, $\lambda_P$, $\lambda_{D_2}$, $\lambda_{L_2}$, which control the weights of different losses, as in traditional regularization approaches \cite{zheng2018regularized,zheng2018minimal}. The output of this algorithm is the trained neural network parameters $\theta_G$, $\theta_D$, $\theta_Q$, $\theta_P$, $\theta_{D_2}$ for network $G$, $D$, $Q$, $P$, $D_2$ respectively. Step 1 initializes network parameters. Then we run the mini-batch loop until $L_1$, $L_2$ and $L_3$ converge. In each mini-batch loop, Step 3 first randomly chooses a batch of real data-label pairs. Step 4 generates batch size of latent code $\cc$ and $\zz$. Step 5 to 9 update all 3 modules.

\begin{figure}[t]
\centering
\begin{subfigure}[b]{0.25\textwidth}
        \includegraphics[width=\textwidth]{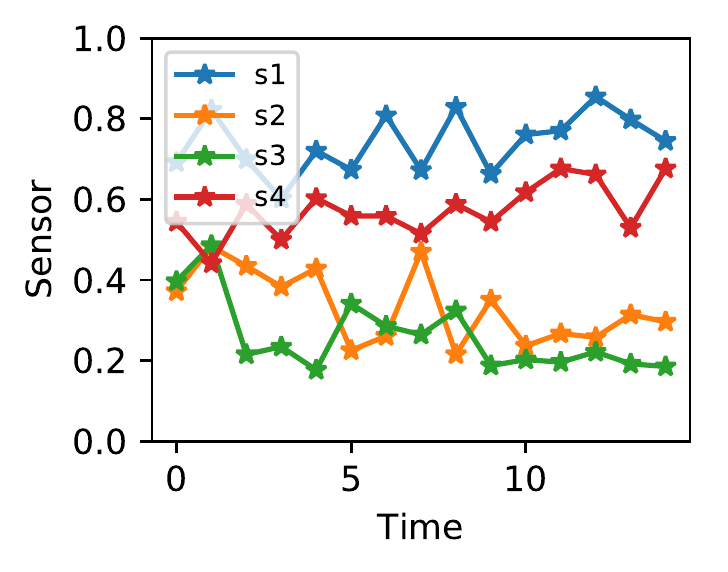}
        \caption{Failure.}
\end{subfigure}%
\begin{subfigure}[b]{0.25\textwidth}
        \includegraphics[width=\textwidth]{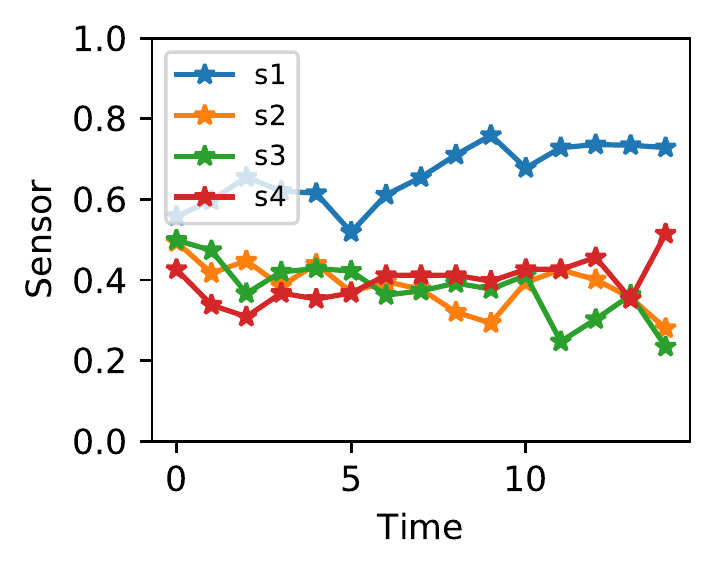}
        \caption{Failure.}
\end{subfigure}%
\begin{subfigure}[b]{0.25\textwidth}
        \includegraphics[width=\textwidth]{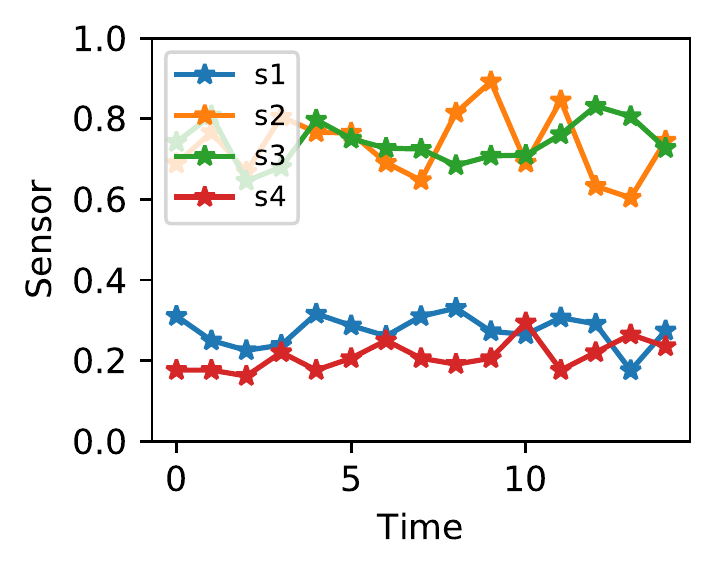}
        \caption{Non-failure.}
\end{subfigure}%
\begin{subfigure}[b]{0.25\textwidth}
        \includegraphics[width=\textwidth]{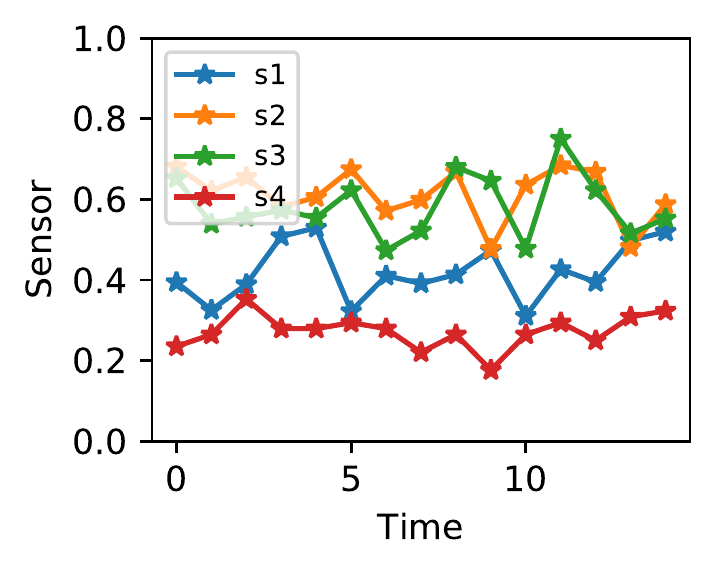}
        \caption{Non-failure.}
\end{subfigure}
\caption{Real CMAPSS FD001 failure and non-failure samples.}
\label{fig:real_sample}
\vspace{0.5cm}
\centering
\begin{subfigure}[b]{0.25\textwidth}
        \includegraphics[width=\textwidth]{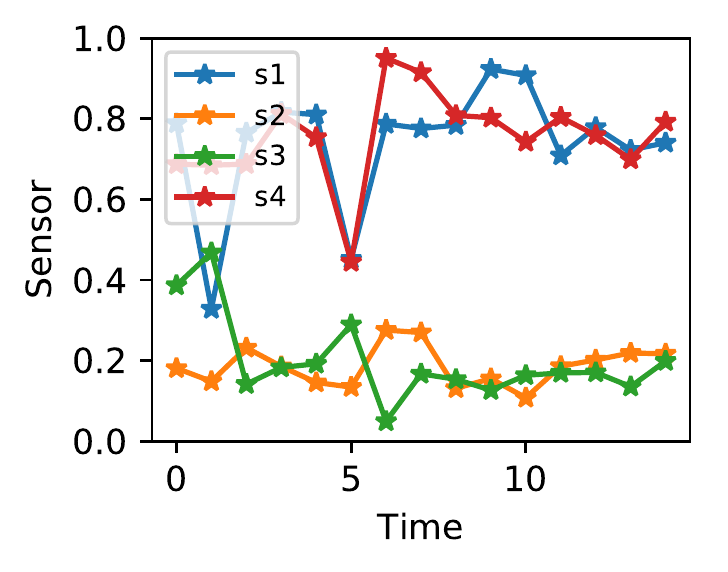}
        \caption{Failure.}
\end{subfigure}%
\begin{subfigure}[b]{0.25\textwidth}
        \includegraphics[width=\textwidth]{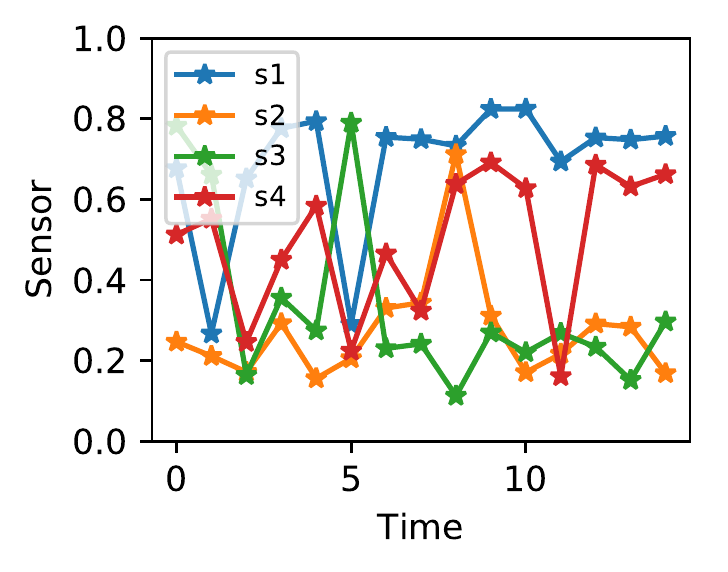}
        \caption{Failure.}
        \label{fig:generate_f2}
\end{subfigure}%
\begin{subfigure}[b]{0.25\textwidth}
        \includegraphics[width=\textwidth]{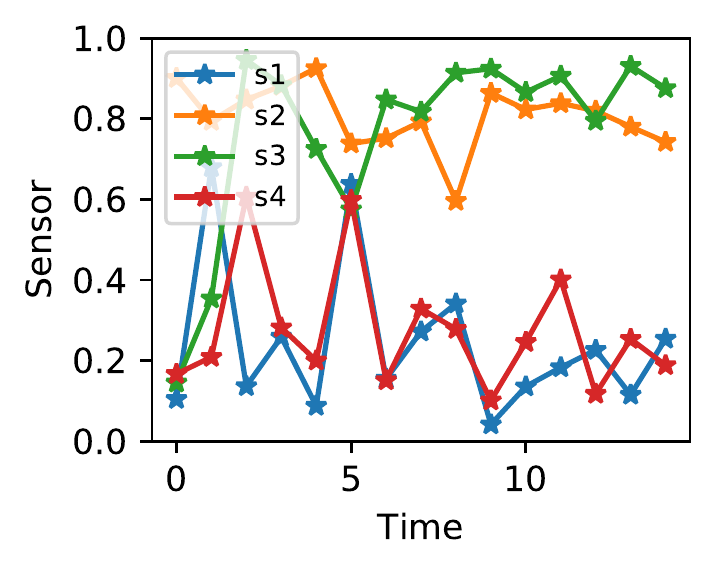}
        \caption{Non-failure.}
\end{subfigure}%
\begin{subfigure}[b]{0.25\textwidth}
        \includegraphics[width=\textwidth]{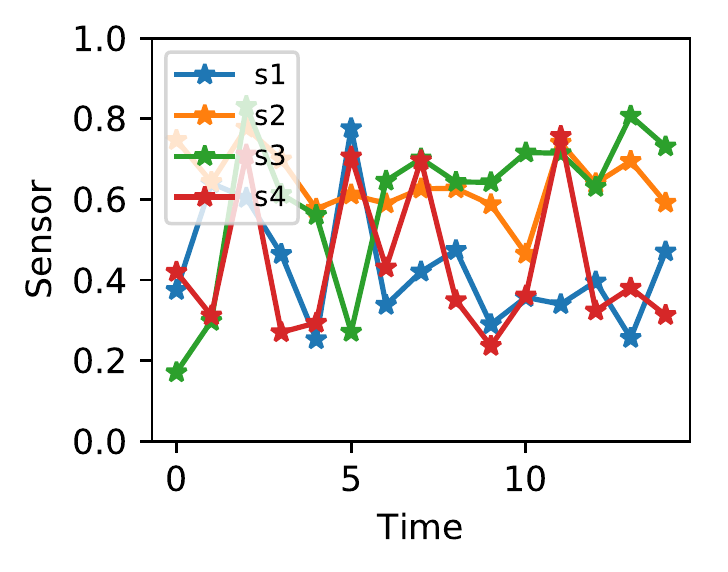}
        \caption{Non-failure.}
        \label{fig:generate_nf2}
\end{subfigure}
\caption{Generated CMAPSS FD001 failure and non-failure samples.}
\label{fig:generated_sample}
\end{figure}

\section{Visualization of generated samples}

We take CMAPSS FD001 data as an example to visualize the generated samples and examine if the proposed GAN-FP can generate realistic enough fake samples for failure prediction task.
CMAPSS FD001 data contains failure and non-failure data for turbofan engines. Each engine sample includes 21 sensors and their readings are in a continuous time window with 15 time steps. Detailed description of CMAPSS FD001 is given in Section \ref{sec:data}. Due to space limitations, we chose 4 sensors (s1, s2, s3, s4) and plotted them from real samples and generated samples. In Figure \ref{fig:real_sample}, we visualize 2 real failure samples and 2 real non-failure samples. As we can see, for failure samples, sensor s1 has higher values than other sensors, sensor s2 and s3 have lower values than s1 and s4, especially for time from 6 to 14; for non-failure samples, sensor s4 has lower values, sensor s2 and s3 have higher values than sensor s1 and s4. In Figure \ref{fig:generated_sample}, we visualize the same 4 sensors from 2 generated failure samples and 2 generated non-failure samples. We observe similar visual properties as in Figure \ref{fig:real_sample}: for failure samples, sensor s2 and s3 have lower values than s1 and s4, especially for time from 6 to 14; for non-failure samples, sensor s2 and s3 have higher values than sensor s1 and s4. We also observe that noises exist in generated samples. For example, at time 11, sensor s4 in Figure \ref{fig:generate_f2} has a big drop, but at time 12, s4 increases back to a higher value. Though real sensor data seems more smooth than generated sensor data, GAN-FP is able to capture the major properties for this failure prediction task. This shows that GAN-FP can generate very good failure and non-failure samples and different levels of variations exist in the generated samples.

\section{Experiments}

\subsection{Data}
\label{sec:data}
We conduct experiments on one Air Pressure System (APS) data set from trucks \cite{apsdata} and four turbofan engine degradation data sets from NASA CMAPSS (Commercial Modular Aero-Propulsion System Simulation) \cite{saxena2008damage}. For APS data, air pressure system generates pressured air that are utilized in various functions in a truck, such as braking and gear changes. The failure class consists of component failures for a specific component of the APS system. The non-failure class consists of samples not related to APS failures. The CMAPSS data consists of four subsets: FD001, FD002, FD003, and FD0004. Data attributes are summarized in Table \ref{tab:data}. In each subset, the data records a snapshot of turbofan engine sensor data at each time cycle, which includes 26 columns: 1st column represents engine ID, 2nd column represents the current operational cycle number, 3-5 columns are the three operational settings that have a substantial effect on engine performance, 6-26 columns represent the 21 sensor values. The engine is operating normally at the start of each time series, and develops a fault at some point in time. The fault grows in magnitude until a system failure. The four CMAPSS data sets have different number of operating conditions and fault conditions. For example, FD001 data has one operating condition and one fault condition, and FD002 has six operating conditions and one fault condition. CMAPSS data set is considered benchmark for predictive maintenance \cite{zheng2017long}.

\begin{table}[t]
\centering
\caption{Data sets.}
\label{tab:data}
\resizebox{\textwidth}{!}{
\begin{tabular}{l||c|c|c|c|c}
\hline\hline
Name & Dimension & Failure sample \# & Non-failure sample \#  & Operating condition \# & Fault condition \#\\
\hline
APS & 170 & 1,000 & 59,000 & N/A & N/A \\
CMAPSS FD001 & 315 & 2,000 & 12,031 & 1 & 1 \\
CMAPSS FD002 & 315 & 5,200 & 31,399 & 6 & 1 \\
CMAPSS FD003 & 315 & 2,000 & 16,210 & 1 & 2 \\
CMAPSS FD004 & 315 & 4,980 & 39,835 & 6 & 2\\
\hline\hline
\end{tabular}
}
\centering
\caption{Network structures.}
\label{tab:network}
\resizebox{0.45\textwidth}{!}{
\begin{tabular}{l||c|c}
\hline\hline
Network & APS & CMAPSS \\
\hline
G & $64,64,170$ & $64,256,500,500,315$ \\
D & $170,64,1$ & $315,500,500,256,1$  \\
Q & $170,64,64,1$ & $315,500,500,256,64,1$  \\
P & $170,64,64,1$ & $315,500,500,256,64,1$  \\
$D_2$ & $171,64,1$ & $316,500,500,256,1$ \\
\hline\hline
\end{tabular}
}
\end{table}

\subsection{Experimental setup and evaluation criteria}
We use fully connected layers for all the networks. Table \ref{tab:network} shows the network structures for both APS and CMAPSS data. For example, G network for APS data consists of 3 layers, with the first layer 64 nodes, second layer 64 nodes and last layer 170 nodes. For APS data, network Q and P share the first two layers with network D. For CMAPSS data, network Q and P share the first four layers with network D. For both APS and CMAPSS data, the noise vector $\zz$ is a $60$-dimensional vector with uniform random values within $[0,1]$. Latent code $\cc$ includes $1$-dimensional categorical code and $3$-dimensional continuous code with uniform random values within $[0,1]$. The activation function is rectified linear unit by default. 

For evaluation, we use AUC (Area Under Curve), (precision, recall, F1) with macro average, micro average, and for the failure class only. All compared methods output the probability that a sample is a failure sample. We then compute the precision and recall curve and calculate AUC. We then can compute both failure and non-failure class precision, recall and F1. Larger values indicate better performance in all these metrics. More about these metrics can be found in \cite{han2011data}.

We compare GAN-FP with 17 other methods. For the first 16 methods, we conduct experiments using 4 classifiers in 4 different sampling settings. The 4 classifiers are DNN, SVM (Support Vector Machines), RF (Random Forests) and DT (Decision Trees). The 4 sampling settings are: undersampling, weighted loss, SMOTE oversampling and ADASYN oversampling. The structure of DNN is the same as network $P$ in GAN-FP. Parameters of SVM, RF and DT are tuned to achieve the best accuracy. For undersampling, we fix the failure samples and randomly draw equal size number of non-failure samples for training. For each experiment, we perform 10 times random undersampling. For weighted loss objective, we assign the same class weights as used in Eq.(\ref{eq:md2}). Lastly, we compare with infoGAN augmented DNN (denoted as InfoGAN AUG), which uses infoGAN to generate more failure samples to make the class distribution balanced and then train a DNN for classification. InfoGAN AUG is used to validate the effectiveness of Module 3. Experiments were performed in a 5-fold cross-validation fashion.

\begin{figure}[h!]
\centering
\begin{minipage}{1\textwidth}
\centering
%\begin{figure}[t]
%\centering
\begin{subfigure}[b]{0.33\textwidth}
        \includegraphics[width=\textwidth]{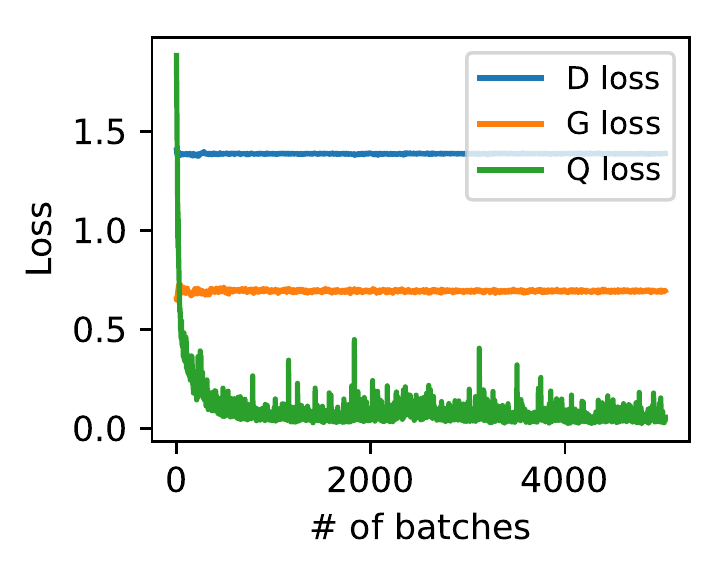}
        \caption{Module 1 loss.}
        \label{fig:loss1}
\end{subfigure}%
\begin{subfigure}[b]{0.33\textwidth}
        \includegraphics[width=\textwidth]{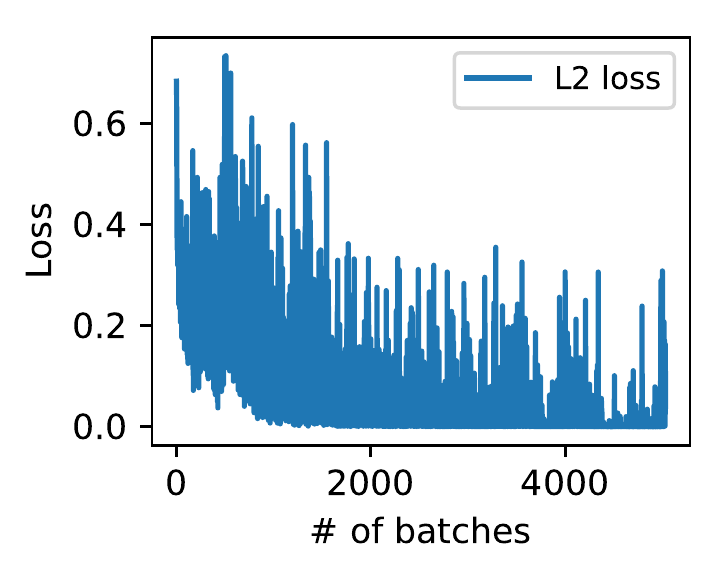}
        \caption{Module 2 loss.}
        \label{fig:loss2}
\end{subfigure}%
\begin{subfigure}[b]{0.33\textwidth}
        \includegraphics[width=\textwidth]{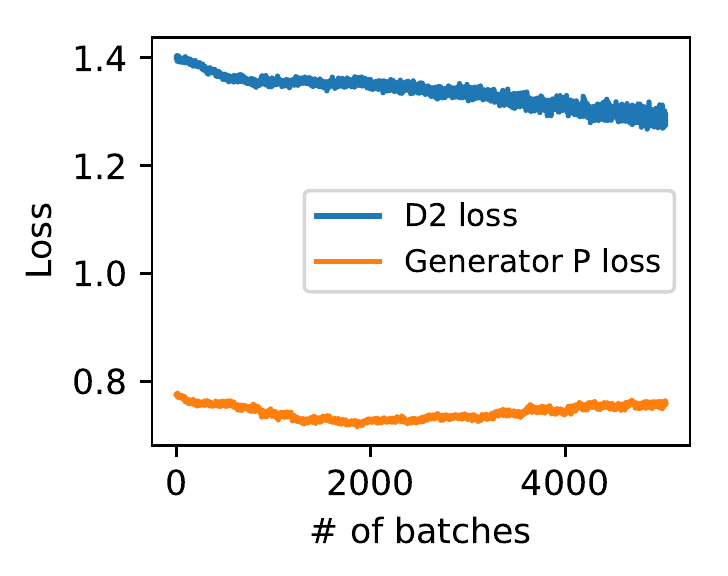}
        \caption{Module 3 loss.}
        \label{fig:loss3}
\end{subfigure}
\caption{Loss.}
%\end{figure}
\end{minipage}
%\begin{figure}[t]
%\centering
\begin{minipage}{1\textwidth}
\vspace{0.5cm}
\centering
\begin{subfigure}[b]{0.235\textwidth}
        \includegraphics[width=\textwidth,height=0.8\textwidth]{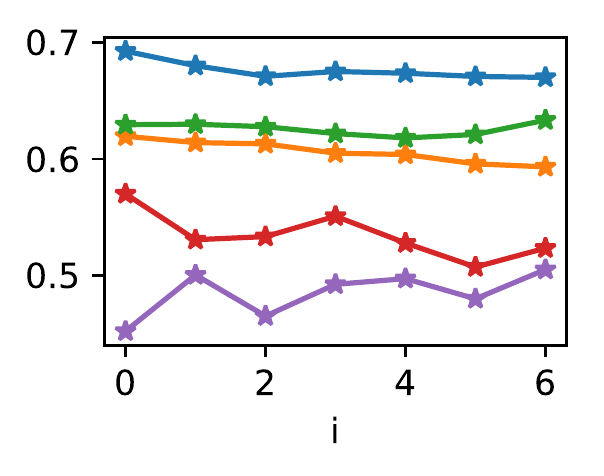}
        \caption{AUC.}
\end{subfigure}%
\begin{subfigure}[b]{0.235\textwidth}
        \includegraphics[width=\textwidth,height=0.8\textwidth]{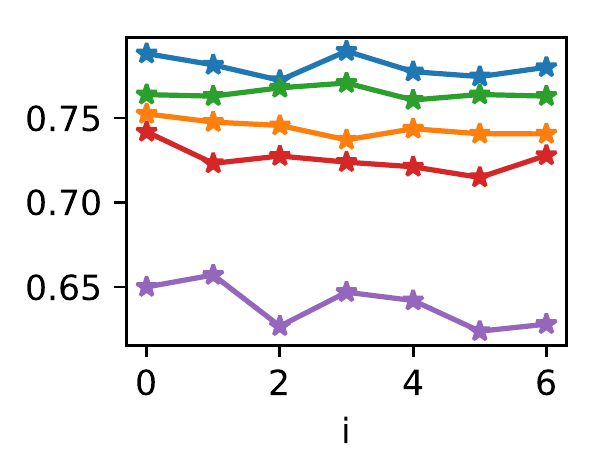}
        \caption{Macro F1.}
\end{subfigure}%
\begin{subfigure}[b]{0.235\textwidth}
        \includegraphics[width=\textwidth,height=0.8\textwidth]{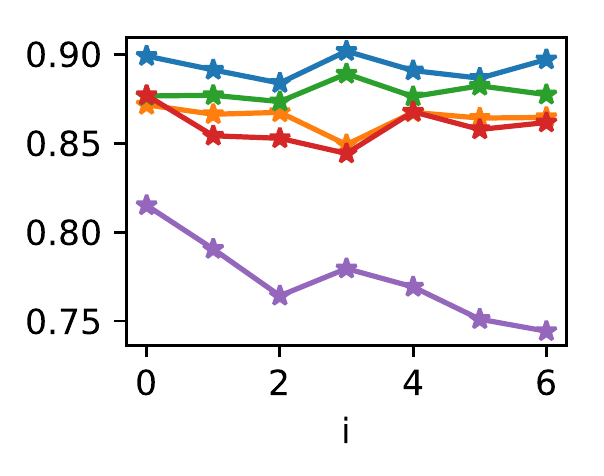}
        \caption{Micro F1.}
\end{subfigure}%
\begin{subfigure}[b]{0.293\textwidth}
        \includegraphics[width=\textwidth,height=0.63\textwidth]{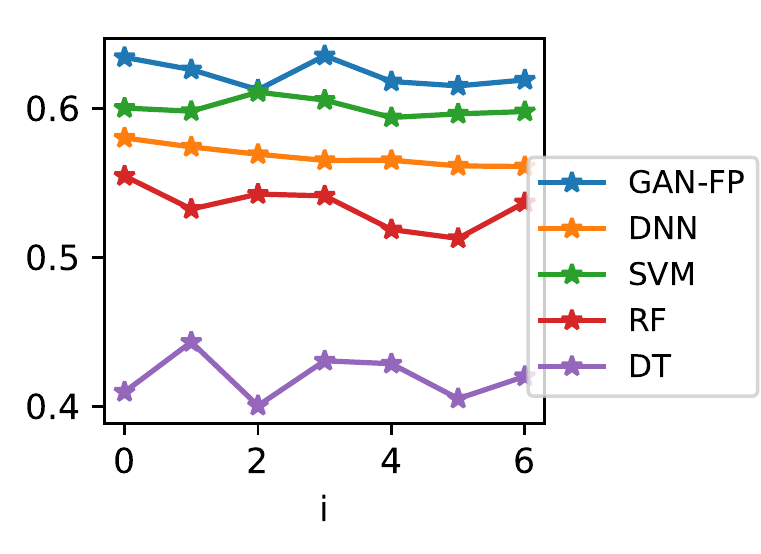}
        \caption{Failure F1.}
\end{subfigure}
\caption{CMAPSS FD001 class imbalance effect using GAN-FP and classifiers with SMOTE: in each figure, the x-axis $i$ indicates $(1000*i)$ non-failure samples are randomly removed from the training data, we do not remove failure samples. The testing samples are fixed for all experiments.}
\label{fig:rate_cmapss}
%\end{figure}
\end{minipage}
\begin{minipage}{1\textwidth}
\vspace{0.5cm}
\centering
\captionsetup{type=table}
%\begin{table}[!]
%\centering
\caption{APS result.}
\label{tab:aps}
\resizebox{0.9\textwidth}{!}{
\begin{tabular}{l|l|l|ccc|ccc|ccc}
\hline\hline
                     &               & \multirow{2}{*}{AUC} & \multicolumn{3}{c|}{Macro}                           & \multicolumn{3}{c|}{Micro}                           & \multicolumn{3}{c}{Failure}                         \\
                     \cline{4-12}
                     &               &                      & Precision       & Recall          & F1              & Precision       & Recall          & F1              & Precision       & Recall          & F1              \\
\hline
\multirow{4}{*}{DNN} & Undersampling & 0.5751               & 0.7393          & 0.8118          & 0.7705          & 0.9827          & 0.9827          & 0.9827          & 0.4847          & 0.6350          & 0.5498          \\
                     & Weighted loss & 0.6131               & 0.8027          & 0.8042          & 0.8034          & 0.9871          & 0.9871          & 0.9871          & 0.6119          & 0.6150          & 0.6135          \\
                     & SMOTE         & 0.7077               & 0.8434          & 0.8350          & 0.8391          & 0.9896          & 0.9896          & 0.9896          & 0.6923          & 0.6750          & 0.6835          \\
                     & ADASYN        & 0.6971               & 0.8040          & 0.8561          & 0.8279          & 0.9878          & 0.9878          & 0.9878          & 0.6128          & 0.7200          & 0.6621          \\
\hline
\multirow{4}{*}{SVM} & Undersampling & 0.3130               & 0.6995          & 0.7706          & 0.7293          & 0.9791          & 0.9791          & 0.9791          & 0.4066          & 0.5550          & 0.4693          \\
                     & Weighted loss & 0.3004               & 0.6829          & 0.7623          & 0.7151          & 0.9773          & 0.9773          & 0.9773          & 0.3737          & 0.5400          & 0.4417          \\
                     & SMOTE         & 0.5673               & 0.7432          & 0.8169          & 0.7749          & 0.9830          & 0.9830          & 0.9830          & 0.4924          & 0.6450          & 0.5584          \\
                     & ADASYN        & 0.5188               & 0.7225          & 0.8158          & 0.7606          & 0.9810          & 0.9810          & 0.9810          & 0.4510          & 0.6450          & 0.5309          \\
\hline
\multirow{4}{*}{RF}  & Undersampling & 0.4274               & 0.6449          & 0.8813          & 0.7052          & 0.9647          & 0.9647          & 0.9647          & 0.2934          & 0.7950          & 0.4286          \\
                     & Weighted loss & 0.3750               & 0.6838          & 0.7333          & 0.7054          & 0.9781          & 0.9781          & 0.9781          & 0.3765          & 0.4800          & 0.4220          \\
                     & SMOTE         & 0.4137               & 0.6602          & 0.7414          & 0.6919          & 0.9747          & 0.9747          & 0.9747          & 0.3289          & 0.5000          & 0.3968          \\
                     & ADASYN        & 0.3387               & 0.6302          & 0.8360          & 0.6832          & 0.9626          & 0.9626          & 0.9626          & 0.2655          & 0.7050          & 0.3858          \\
\hline
\multirow{4}{*}{DT}  & Undersampling & 0.5614               & 0.5928          & \textbf{0.9330}          & 0.6376          & 0.9311          & 0.9311          & 0.9311          & 0.1868          & \textbf{0.9350}          & 0.3114          \\
                     & Weighted loss & 0.6310               & 0.8194          & 0.8022          & 0.8106          & 0.9879          & 0.9879          & 0.9879          & 0.6455          & 0.6100          & 0.6272          \\
                     & SMOTE         & 0.6471               & 0.7751          & 0.8625          & 0.8125          & 0.9858          & 0.9858          & 0.9858          & 0.5547          & 0.7350          & 0.6323          \\
                     & ADASYN        & 0.6094               & 0.7567          & 0.8420          & 0.7930          & 0.9842          & 0.9842          & 0.9842          & 0.5187          & 0.6950          & 0.5940          \\
\hline
\multicolumn{2}{l|}{InfoGAN AUG}      & 0.7343               & 0.8335          & 0.8744          & 0.8527          & 0.9898          & 0.9898          & 0.9898          & 0.6711          & 0.7550          & 0.7106          \\
\hline
\multicolumn{2}{l|}{GAN-FP}            & \textbf{0.8085}      & \textbf{0.8662} & 0.8955 & \textbf{0.8803} & \textbf{0.9918} & \textbf{0.9918} & \textbf{0.9918} & \textbf{0.7358} & 0.7959 & \textbf{0.7647} \\
\hline\hline
\end{tabular}
}
%\end{table}
\end{minipage}
\end{figure}

\subsection{Algorithm convergence}
To examine the training convergence, we take CMAPSS FD001 data as an example and plot the loss changes along the training process. From Eq.(\ref{eq:infogan}), we know that Module 1 loss consists of three parts: discriminator (D) loss $\min_D -V(D)$, generator (G) loss $\min_{G} V(G)$ and mutual information (Q) loss $\min_{G,Q} - L_{mutual}(G,Q)$. 
Figure \ref{fig:loss1} shows that D loss and G loss converge along the training. Mutual information (Q) loss is minimized and converged after about 2,000 batches. Advanced accelerating algorihtm can reduce training time furthermore \cite{zheng2016accelerating}. Figure \ref{fig:loss2} shows that $L_2$ loss Eq.(\ref{eq:md2}) is decreasing from 0 to 3,000 batches. Figure \ref{fig:loss3} shows that D2 loss and generator P loss of Module 3 converge along the training. Overall, this shows the effectiveness of Algorithm \ref{alg:alg1}.

\subsection{Effect of class imbalance}
We compare the classification performance of different approaches when the number of majority non-failure samples is decreased. Figure \ref{fig:rate_cmapss} shows that, for CMAPSS FD001, there is no significant performance loss when the number of majority non-failure samples is decreased. Among all experiments, GAN-FP gives the best performance in terms of the four metrics.

\begin{table}[t!]
\centering
\caption{CMAPSS FD001.}
\label{tab:fd001}
\resizebox{1.0\textwidth}{!}{
\begin{tabular}{l|l|l|ccc|ccc|ccc}
\hline\hline
                     &               & \multirow{2}{*}{AUC} & \multicolumn{3}{c|}{Macro}                           & \multicolumn{3}{c|}{Micro}                           & \multicolumn{3}{c}{Failure}                         \\
                     \cline{4-12}
                     &               &                      & Precision       & Recall          & F1              & Precision       & Recall          & F1              & Precision       & Recall          & F1              \\
\hline
\multirow{4}{*}{DNN} & Undersampling & 0.6381               & 0.7525          & 0.7895          & 0.7687          & 0.8785          & 0.8785          & 0.8785          & 0.5624          & 0.6650          & 0.6094          \\
                     & Weighted loss & 0.6030               & 0.7327          & 0.7614          & 0.7455          & 0.8678          & 0.8678          & 0.8678          & 0.5315          & 0.6125          & 0.5691          \\
                     & SMOTE         & 0.6196               & 0.7397          & 0.7678          & 0.7524          & 0.8717          & 0.8717          & 0.8717          & 0.5437          & 0.6225          & 0.5804          \\
                     & ADASYN        & 0.6185               & 0.7473          & 0.7559          & 0.7515          & 0.8764          & 0.8764          & 0.8764          & 0.5635          & 0.5875          & 0.5753          \\
\hline
\multirow{4}{*}{SVM} & Undersampling & 0.6331               & 0.7592          & 0.7720          & 0.7653          & 0.8824          & 0.8824          & 0.8824          & 0.5825          & 0.6175          & 0.5995          \\
                     & Weighted loss & 0.6498               & 0.7485          & \textbf{0.7972} & 0.7689          & 0.8757          & 0.8757          & 0.8757          & 0.5511          & 0.6875          & 0.6118          \\
                     & SMOTE         & 0.6295               & 0.7491          & 0.7822          & 0.7638          & 0.8767          & 0.8767          & 0.8767          & 0.5579          & 0.6500          & 0.6005          \\
                     & ADASYN        & 0.6224               & 0.7461          & 0.7893          & 0.7646          & 0.8746          & 0.8746          & 0.8746          & 0.5492          & 0.6700          & 0.6036          \\
\hline
\multirow{4}{*}{RF}  & Undersampling & 0.5531               & 0.7046          & 0.7466          & 0.7218          & 0.8497          & 0.8497          & 0.8497          & 0.4782          & 0.6025          & 0.5332          \\
                     & Weighted loss & 0.5378               & 0.7067          & 0.7504          & 0.7245          & 0.8507          & 0.8507          & 0.8507          & 0.4813          & 0.6100          & 0.5380          \\
                     & SMOTE         & 0.5701               & 0.7486          & 0.7355          & 0.7418          & 0.8771          & 0.8771          & 0.8771          & 0.5733          & 0.5375          & 0.5548          \\
                     & ADASYN        & 0.5238               & 0.7322          & 0.7134          & 0.7221          & 0.8696          & 0.8696          & 0.8696          & 0.5470          & 0.4950          & 0.5197          \\
\hline
\multirow{4}{*}{DT}  & Undersampling & 0.5279               & 0.6096          & 0.7109          & 0.5977          & 0.6901          & 0.6901          & 0.6901          & 0.2787          & \textbf{0.7400} & 0.4049          \\
                     & Weighted loss & 0.4699               & 0.6631          & 0.6695          & 0.6662          & 0.8336          & 0.8336          & 0.8336          & 0.4200          & 0.4400          & 0.4298          \\
                     & SMOTE         & 0.4521               & 0.6406          & 0.6629          & 0.6500          & 0.8151          & 0.8151          & 0.8151          & 0.3758          & 0.4500          & 0.4096          \\
                     & ADASYN        & 0.4471               & 0.6373          & 0.6596          & 0.6466          & 0.8130          & 0.8130          & 0.8130          & 0.3701          & 0.4450          & 0.4041          \\
\hline
\multicolumn{2}{l|}{InfoGAN AUG}      & 0.6128               & 0.7256          & 0.7716          & 0.7446          & 0.8621          & 0.8621          & 0.8621          & 0.5129          & 0.6450          & 0.5714          \\
\hline
\multicolumn{2}{l|}{GAN-FP}            & \textbf{0.6927}      & \textbf{0.8021} & 0.7759          & \textbf{0.7881} & \textbf{0.8992} & \textbf{0.8992} & \textbf{0.8992} & \textbf{0.6707} & 0.6022          & \textbf{0.6346} \\
\hline\hline
\end{tabular}
}
\centering
\caption{CMAPSS FD002.}
\label{tab:fd002}
\resizebox{1.0\textwidth}{!}{
\begin{tabular}{l|l|l|ccc|ccc|ccc}
\hline\hline
                     &               & \multirow{2}{*}{AUC} & \multicolumn{3}{c|}{Macro}                           & \multicolumn{3}{c|}{Micro}                           & \multicolumn{3}{c}{Failure}                         \\
                     \cline{4-12}
                     &               &                      & Precision       & Recall          & F1              & Precision       & Recall          & F1              & Precision       & Recall          & F1              \\
\hline
\multirow{4}{*}{DNN} & Undersampling & 0.5503               & 0.6996          & 0.7501          & 0.7193          & 0.8452    & 0.8452 & 0.8452 & 0.4662    & 0.6173          & 0.5312          \\ 
                     & Weighted loss & 0.5431               & 0.7014          & 0.7549          & 0.7219          & 0.8458          & 0.8458          & 0.8458          & 0.4681          & 0.6279          & 0.5363          \\
                     & SMOTE         & 0.5383               & 0.6920          & \textbf{0.7675} & 0.7166          & 0.8331          & 0.8331          & 0.8331          & 0.4427          & 0.6760          & 0.5350          \\
                     & ADASYN        & 0.5377               & 0.6910          & 0.7666          & 0.7156          & 0.8322          & 0.8322          & 0.8322          & 0.4410          & 0.6750          & 0.5334          \\
\hline
\multirow{4}{*}{SVM} & Undersampling & 0.5212               & 0.6888          & 0.7057          & 0.6965          & 0.8454          & 0.8454          & 0.8454          & 0.4601          & 0.5106          & 0.4840          \\
                     & Weighted loss & 0.5199               & 0.6869          & 0.7021          & 0.6939          & 0.8447          & 0.8447          & 0.8447          & 0.4576          & 0.5029          & 0.4792          \\
                     & SMOTE         & 0.5438               & 0.6844          & 0.7435          & 0.7055          & 0.8324          & 0.8324          & 0.8324          & 0.4366          & 0.6192          & 0.5121          \\
                     & ADASYN        & 0.5444               & 0.6917          & 0.7334          & 0.7085          & 0.8419          & 0.8419          & 0.8419          & 0.4559          & 0.5817          & 0.5112          \\
\hline
\multirow{4}{*}{RF}  & Undersampling & 0.4610               & 0.6649          & 0.7293          & 0.6853          & 0.8149          & 0.8149          & 0.8149          & 0.4005          & 0.6096          & 0.4834          \\
                     & Weighted loss & 0.4836               & 0.6675          & 0.7234          & 0.6868          & 0.8206          & 0.8206          & 0.8206          & 0.4087          & 0.5875          & 0.4821          \\
                     & SMOTE         & 0.4368               & 0.6519          & 0.7225          & 0.6714          & 0.7999          & 0.7999          & 0.7999          & 0.3752          & 0.6144          & 0.4659          \\
                     & ADASYN        & 0.4409               & 0.6524          & 0.7204          & 0.6718          & 0.8018          & 0.8018          & 0.8018          & 0.3772          & 0.6067          & 0.4652          \\
\hline
\multirow{4}{*}{DT}  & Undersampling & 0.4956               & 0.5892          & 0.6753          & 0.5668          & 0.6583          & 0.6583          & 0.6583          & 0.2494          & \textbf{0.6990} & 0.3676          \\
                     & Weighted loss & 0.4149               & 0.6306          & 0.6338          & 0.6322          & 0.8184          & 0.8184          & 0.8184          & 0.3651          & 0.3760          & 0.3704          \\
                     & SMOTE         & 0.3949               & 0.5848          & 0.6221          & 0.5929          & 0.7549          & 0.7549          & 0.7549          & 0.2732          & 0.4365          & 0.3360          \\
                     & ADASYN        & 0.4244               & 0.5999          & 0.6451          & 0.6104          & 0.7641          & 0.7641          & 0.7641          & 0.2959          & 0.4788          & 0.3658          \\
\hline
\multicolumn{2}{l|}{InfoGAN AUG}      & 0.5484               & 0.6945          & 0.7658          & 0.7187          & 0.8363          & 0.8363          & 0.8363          & 0.4489          & 0.6673          & 0.5367          \\
\hline
\multicolumn{2}{l|}{GAN-FP}            & \textbf{0.5666}      & \textbf{0.7081} & 0.7488          & \textbf{0.7249} & \textbf{0.8521}          & \textbf{0.8521}          & \textbf{0.8521}          & \textbf{0.4847}          & 0.6043          & \textbf{0.5379}
\\
\hline\hline
\end{tabular}
}
\end{table}

\begin{table}[t!]
\centering
\caption{CMAPSS FD003.}
\label{tab:fd003}
\resizebox{1.0\textwidth}{!}{
\begin{tabular}{l|l|l|ccc|ccc|ccc}
\hline\hline
                     &               & \multirow{2}{*}{AUC} & \multicolumn{3}{c|}{Macro}                           & \multicolumn{3}{c|}{Micro}                           & \multicolumn{3}{c}{Failure}                         \\
                     \cline{4-12}
                     &               &                      & Precision       & Recall          & F1              & Precision       & Recall          & F1              & Precision       & Recall          & F1              \\
\hline
\multirow{4}{*}{DNN} & Undersampling & 0.6211               & 0.7423          & 0.7998          & 0.7663          & 0.8971          & 0.8971          & 0.8971          & 0.5263          & 0.6750          & 0.5915          \\
                     & Weighted loss & 0.6035               & \textbf{0.7654} & 0.7654          & 0.7654          & \textbf{0.9078} & \textbf{0.9078} & \textbf{0.9078} & \textbf{0.5825} & 0.5825          & 0.5825          \\
                     & SMOTE         & 0.6207               & 0.7376          & 0.8142          & 0.7674          & 0.8935          & 0.8935          & 0.8935          & 0.5126          & 0.7125          & 0.5962          \\
                     & ADASYN        & 0.6199               & 0.7451          & 0.7964          & 0.7670          & 0.8987          & 0.8987          & 0.8987          & 0.5331          & 0.6650          & 0.5918          \\
\hline
\multirow{4}{*}{SVM} & Undersampling & 0.5718               & 0.7469          & 0.7666          & 0.7562          & 0.9004          & 0.9004          & 0.9004          & 0.5446          & 0.5950          & 0.5687          \\
                     & Weighted loss & 0.6338               & 0.7449          & 0.8123          & 0.7722          & 0.8979          & 0.8979          & 0.8979          & 0.5282          & 0.7025          & 0.6030          \\
                     & SMOTE         & 0.6166               & 0.7356          & 0.7967          & 0.7606          & 0.8935          & 0.8935          & 0.8935          & 0.5134          & 0.6725          & 0.5823          \\
                     & ADASYN        & 0.6113               & 0.7480          & 0.7799          & 0.7625          & 0.9007          & 0.9007          & 0.9007          & 0.5435          & 0.6250          & 0.5814          \\
\hline
\multirow{4}{*}{RF}  & Undersampling & 0.5152               & 0.7223          & 0.7778          & 0.7452          & 0.8871          & 0.8871          & 0.8871          & 0.4913          & 0.6375          & 0.5550          \\
                     & Weighted loss & 0.5693               & 0.7520          & 0.7602          & 0.7560          & 0.9026          & 0.9026          & 0.9026          & 0.5566          & 0.5775          & 0.5669          \\
                     & SMOTE         & 0.5266               & 0.7164          & 0.7933          & 0.7454          & 0.8816          & 0.8816          & 0.8816          & 0.4747          & 0.6800          & 0.5591          \\
                     & ADASYN        & 0.5276               & 0.7120          & 0.7866          & 0.7402          & 0.8794          & 0.8794          & 0.8794          & 0.4676          & 0.6675          & 0.5499          \\
\hline
\multirow{4}{*}{DT}  & Undersampling & 0.5460               & 0.6206          & 0.7671          & 0.6229          & 0.7453          & 0.7453          & 0.7453          & 0.2744          & 0.7950          & 0.4080          \\
                     & Weighted loss & 0.4309               & 0.6625          & 0.6607          & 0.6616          & 0.8678          & 0.8678          & 0.8678          & 0.4000          & 0.3950          & 0.3975          \\
                     & SMOTE         & 0.4751               & 0.6604          & 0.7052          & 0.6780          & 0.8554          & 0.8554          & 0.8554          & 0.3839          & 0.5125          & 0.4390          \\
                     & ADASYN        & 0.4364               & 0.6408          & 0.6782          & 0.6555          & 0.8463          & 0.8463          & 0.8463          & 0.3510          & 0.4625          & 0.3991          \\
\hline
\multicolumn{2}{l|}{InfoGAN AUG}      & 0.6167               & 0.7620          & 0.7768          & 0.7691          & 0.9067          & 0.9067          & 0.9067          & 0.5728          & 0.6100          & 0.5908          \\
\hline
\multicolumn{2}{l|}{GAN-FP}            & \textbf{0.7093}      & 0.7584          & \textbf{0.8635} & \textbf{0.7970} & 0.9040          & 0.9040          & 0.9040          & 0.5416          & \textbf{0.8117} & \textbf{0.6497}
\\
\hline\hline
\end{tabular}
}
\centering
\caption{CMAPSS FD004.}
\label{tab:fd004}
\resizebox{1.0\textwidth}{!}{
\begin{tabular}{l|l|l|ccc|ccc|ccc}
\hline\hline
                     &               & \multirow{2}{*}{AUC} & \multicolumn{3}{c|}{Macro}                           & \multicolumn{3}{c|}{Micro}                           & \multicolumn{3}{c}{Failure}                         \\
                     \cline{4-12}
                     &               &                      & Precision       & Recall          & F1              & Precision       & Recall          & F1              & Precision       & Recall          & F1              \\
\hline
\multirow{4}{*}{DNN} & Undersampling & 0.4826               & 0.7038          & 0.7719          & 0.7297          & 0.8748          & 0.8748          & 0.8748          & 0.4550          & 0.6396          & 0.5317          \\ 
                     & Weighted loss & 0.5065               & 0.7197          & 0.7788          & 0.7436          & 0.8847          & 0.8847          & 0.8847          & 0.4860          & 0.6426          & 0.5534          \\
                     & SMOTE         & 0.5360               & 0.7107          & \textbf{0.7802} & 0.7374          & 0.8786          & 0.8786          & 0.8786          & 0.4670          & 0.6536          & 0.5448          \\  
                     & ADASYN        & 0.5398               & 0.7192          & 0.7711          & 0.7408          & 0.8852          & 0.8852          & 0.8852          & 0.4871          & 0.6245          & 0.5473          \\ \hline
\multirow{4}{*}{SVM} & Undersampling & 0.4588               & 0.6770          & 0.7760          & 0.7070          & 0.8501          & 0.8501          & 0.8501          & 0.3979          & 0.6807          & 0.5022          \\
                     & Weighted loss & 0.4553               & 0.6749          & 0.7756          & 0.7048          & 0.8478          & 0.8478          & 0.8478          & 0.3935          & 0.6827          & 0.4993          \\
                     & SMOTE         & 0.4983               & 0.6986          & 0.7767          & 0.7268          & 0.8700          & 0.8700          & 0.8700          & 0.4428          & 0.6566          & 0.5289          \\ 
                     & ADASYN        & 0.4978               & 0.6979          & 0.7708          & 0.7248          & 0.8706          & 0.8706          & 0.8706          & 0.4432          & 0.6426          & 0.5246          \\ \hline
\multirow{4}{*}{RF}  & Undersampling & 0.4748               & 0.6914          & 0.7335          & 0.7090          & 0.8721          & 0.8721          & 0.8721          & 0.4403          & 0.5552          & 0.4911          \\ 
                     & Weighted loss & 0.4685               & 0.6836          & 0.7421          & 0.7060          & 0.8648          & 0.8648          & 0.8648          & 0.4217          & 0.5843          & 0.4899          \\ 
                     & SMOTE         & 0.4227               & 0.6730          & 0.7612          & 0.7010          & 0.8503          & 0.8503          & 0.8503          & 0.3941          & 0.6466          & 0.4897          \\ 
                     & ADASYN        & 0.4035               & 0.6618          & 0.7493          & 0.6882          & 0.8417          & 0.8417          & 0.8417          & 0.3740          & 0.6305          & 0.4695          \\ \hline
\multirow{4}{*}{DT}  & Undersampling & 0.4947               & 0.5996          & 0.7195          & 0.5957          & 0.7263          & 0.7263          & 0.7263          & 0.2464          & \textbf{0.7108} & 0.3660          \\  
                     & Weighted loss & 0.4026               & 0.6450          & 0.6437          & 0.6443          & 0.8601          & 0.8601          & 0.8601          & 0.3692          & 0.3655          & 0.3673          \\ 
                     & SMOTE         & 0.4243               & 0.6118          & 0.6746          & 0.6285          & 0.8097          & 0.8097          & 0.8097          & 0.2922          & 0.5010          & 0.3691          \\ 
                     & ADASYN        & 0.4195               & 0.6097          & 0.6709          & 0.6259          & 0.8085          & 0.8085          & 0.8085          & 0.2887          & 0.4940          & 0.3644          \\ \hline
\multicolumn{2}{l|}{InfoGAN AUG}    & 0.5581               & 0.7209          & 0.7784          & 0.7443          & 0.8855          & 0.8855          & 0.8855          & 0.4885          & 0.6406          & 0.5543          \\ \hline
\multicolumn{2}{l|}{GAN-FP}          & \textbf{0.5638}      & \textbf{0.7260} & 0.7773          & \textbf{0.7475} & \textbf{0.8890} & \textbf{0.8890} & \textbf{0.8890} & \textbf{0.4992} & 0.6338          & \textbf{0.5585}
\\
\hline\hline
\end{tabular}
}
\end{table}

\subsection{Comparison with other methods}
Table \ref{tab:aps} shows the result for APS data. Table \ref{tab:fd001}, \ref{tab:fd002}, \ref{tab:fd003} and \ref{tab:fd004} show the results for CMAPSS. The best performing methods in each column is in bold. Note that CMAPSS data sets have different levels of difficulty since they have different number of operating conditions and fault conditions. Overall, GAN-FP shows better results in terms of AUC and F1 score compared to its counterparts. The fact that GAN-FP outperforms InfoGAN AUG shows the effectiveness of Module 3.

\section{Conclusion}
In conclusion, we proposed a novel model GAN-FP for imbalanced classification and failure prediction, and experimented it on industrial data. This novel design not only improves modeling performance, but also can have significant potential economical and social values.

\end{document}